\PassOptionsToPackage{table}{xcolor}
\documentclass{applemlr}
\usepackage{amsmath,amssymb}
\usepackage{array,tabularx,makecell}
\newcolumntype{P}[1]{>{\raggedright\arraybackslash}p{#1}}
\usepackage{pifont}
\usepackage{xspace}
\usepackage{enumitem}
\usepackage{fvextra}
\usepackage{needspace}
\definecolor{lightgray}{gray}{0.94}
\newcommand{\racap}{\mbox{RACaP}\xspace}

\hypersetup{
  pdftitle={RACaP: Agentic Reasoning, Acting, and Coding as Policies for Evolvable Robot Learning},
  pdfauthor={Zexi Li; Yehang Zhang; Haojian Huang; Bohan Zhou; Wenqian Li; Chenxu Wang; Yifan Chang; Yangkai Wei; Tianyi Zhang; Ying-Cong Chen; Kaiwen Zhou; Yinchuan Li; James Cheng},
  pdfsubject={Evolvable robot learning},
  pdfkeywords={RACaP, robot learning, Code as Policies, ReAct, agentic evolution},
  bookmarksnumbered=true
}
\renewcommand\authorformat[2][]{{\sffamily\bfseries\mbox{#2$^{#1}$}}}
\renewcommand\affiliationformat[2][]{{\small\sffamily $^{#1}$#2}}
\renewcommand\affiliation[2][]{\addtolist[#1]{#2}{\affiliationlist}{\affiliationformat}{\par}}
\newcommand{\repositorylink}[1]{%
  \href{#1}{{\sffamily\mdseries\fontsize{9}{11}\selectfont
    \texttt{</>}\enspace GitHub}}}
\newcommand{\paperlinks}{\repositorylink{https://github.com/Robo-Harness/racap}}
\patchcmd{\mymaketitle}{\abstractlist\par}
  {\ifdefempty{\paperlinks}{}{\paperlinks\par}\abstractlist\par}
  {}{\ClassError{applemlr}{Cannot insert paper links}{Check the title layout.}}
\title{RACaP: Agentic Reasoning, Acting, and Coding as Policies for Evolvable Robot Learning}
\author[1,3\ast]{Zexi Li}
\author[2,3]{Yehang Zhang}
\author[2,3]{Haojian Huang}
\author[1,3]{Bohan Zhou}
\author[1,3]{Wenqian Li}
\author[3]{Chenxu Wang}
\author[3]{Yifan Chang}
\author[3]{Yangkai Wei}
\author[3]{Tianyi Zhang}
\author[2]{Ying-Cong Chen}
\author[3]{Kaiwen Zhou}
\author[3]{Yinchuan Li}
\author[1\ast]{James Cheng}
\renewcommand{\authorlist}{%
  \authorformat[1,3\ast]{Zexi Li}, \authorformat[2,3]{Yehang Zhang},
  \authorformat[2,3]{Haojian Huang}, \authorformat[1,3]{Bohan Zhou},
  \authorformat[1,3]{Wenqian Li},\\
  \authorformat[3]{Chenxu Wang}, \authorformat[3]{Yifan Chang},
  \authorformat[3]{Yangkai Wei}, \authorformat[3]{Tianyi Zhang},\\
  \authorformat[2]{Ying-Cong Chen}, \authorformat[3]{Kaiwen Zhou},
  \authorformat[3]{Yinchuan Li}, \authorformat[1\ast]{James Cheng}}
\affiliation[1]{The Chinese University of Hong Kong}
\affiliation[2]{The Hong Kong University of Science and Technology (Guangzhou)}
\affiliation[3]{Knowin AI}
\contribution[\ast]{Corresponding authors: \email{zexili@cuhk.edu.hk}; \email{jcheng@cse.cuhk.edu.hk}}

\abstract{General-purpose robot agents must learn from experience, transfer their knowledge to new tasks, and act efficiently during physical execution. Code as Policies (CaP) methods provide flexibility by generating and repairing robot programs after a task arrives. However, runtime coding is slow and often encodes task-specific coordinates and recovery decisions, limiting generalization and transfer. Although recent self-improving CaP methods use experience from past programs, they still rely on runtime code generation, and their ambiguous play-time objectives may lead to experiences that are not applicable to downstream tasks. In this paper, we address the evolution-to-execution problem by moving coding to the evolution stage, while using more flexible and efficient function calls within a Reasoning-and-Acting (ReAct) loop at runtime. To this end, we introduce \racap, an agentic framework that evolves typed policy application programming interfaces (Policy APIs), a ReAct agent harness, and experience memory before deployment. During evolution, we adopt a two-phase strategy that combines capability curriculum learning with autonomous self-evolution, enabling the system to learn and evolve gradually and steadily. During this process, the Policy APIs are optimized for generalizability and steerability. The ReAct agent maintains task-objective-driven working memory and experience-based long-term memory, enabling it to handle long-horizon tasks. At runtime, it uses visual observations and perceptual feedback to call Policy APIs, adapt its strategy, and respond promptly to changes in physical state. Experiments demonstrate the generalization, transferability, and efficiency of RACaP. On LIBERO, RACaP achieves 54.4\% in-domain success, transfers zero-shot to LIBERO-PRO with 45.0\% success, and solves 46.0\% of long-horizon tasks on LIBERO-Long, compared with at most 4.0\% for CaP baselines. On LIBERO-PRO, RACaP achieves $2.5\times$ the success rate of CaP baselines while providing a $1.9\times$ speedup in median policy time. For more efficient on-robot edge deployment, we use rejection-sampled fine-tuning to distill the ReAct agent's decision traces from GPT-5.6 into a compact Qwen3-VL-8B-Instruct model. The distilled model achieves a $13.2\times$ per-decision inference speedup and reduces repeated physical calls from 16 to 4. These results show that separating reusable code from runtime decisions enables continued evolution, effective transfer, and efficient long-horizon control.}

\begin{document}
\maketitle

\section{Introduction}

General-purpose robot agents need to learn from experience and reuse what they learn in new physical tasks~\citep{bousmalis2023robocat}, and they also adapt while a task unfolds~\citep{chen2026adapt,worldlines}. For example, an instruction such as ``put all the utensils on the plate and close the drawer'' requires the robot to identify several objects, order causally dependent actions, verify each physical outcome, and recover from mistakes. The objects and layout may change in the next episode. A useful robot policy must therefore combine knowledge that transfers across tasks with decisions that remain responsive to the~current~scene. \looseness=-1

Code as Policies (CaP) offers a flexible way to make these decisions by generating executable robot programs from language instructions~\citep{code_as_policies}. CaP-X further uses visual feedback to repair a program after execution~\citep{cap_x}. This approach can handle a new task without training a new neural policy, but it places coding on the physical execution path. One coding iteration takes 6.8 to 23.8 seconds in CaP-X, and a multi-turn trial takes 60.8 to 113.6 seconds. Runtime code may also introduce syntax errors, select the wrong object, or repeat an ineffective action while the robot is already operating.

\begin{figure}[tbp]
\centering
\includegraphics[width=0.98\textwidth]{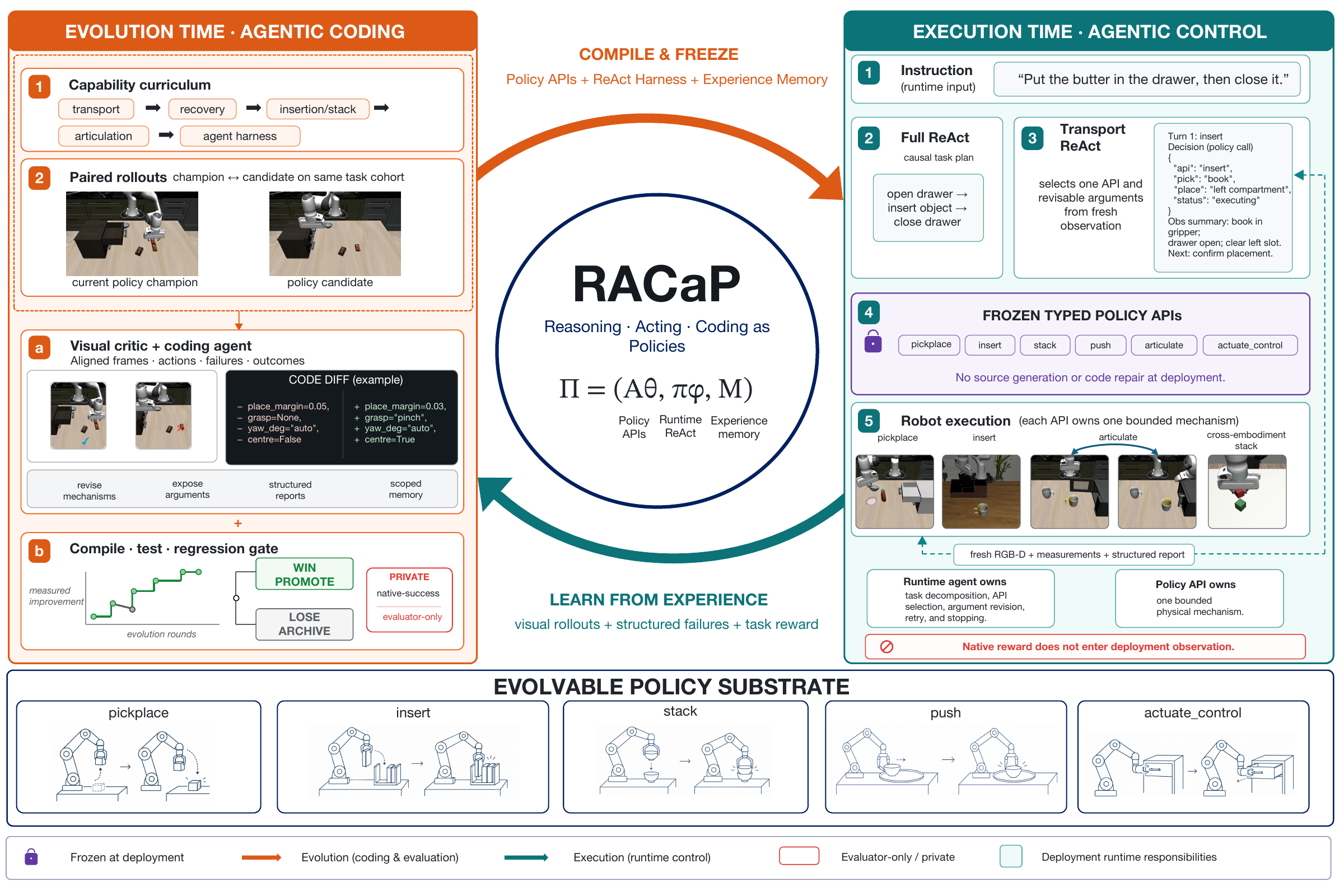}
\caption{\textbf{\racap separates code evolution from runtime execution.} During evolution, a capability curriculum produces paired rollouts, a visual critic and coding agent revise reusable mechanisms, and native success selects the next version. During deployment, a ReAct agent calls frozen typed Policy APIs and changes only their exposed arguments. Evaluator state never enters the deployment observation.}
\label{fig:method}
\end{figure}

Runtime coding cannot be easily transferred across tasks because one generated program often contains reusable physical mechanisms, but also task-specific setups and offsets, such as an object name, a coordinate, or a specific grasp height. Therefore, changing the tasks will require writing new codes from scratch. 
Recent self-improving CaP methods begin to use past execution experience, but they do not fully resolve this conflict. Playful Agentic Robot Learning (RATS) stores programs learned through a play objective based on novelty and learnability~\citep{rats}. However, this objective does not guarantee whether a learned program has a positive effect on downstream tasks. In our controlled study, the play in LIBERO-90 reduces RATS success from 21.1\% to 11.1\% in-domain and from 17.8\% to 8.3\% on zero-shot LIBERO-PRO. In addition, ASPIRE validates and retrieves program repairs~\citep{aspire}, but many repairs remain tied to fixed distances or grasp heights from the observed failure~\citep{aspire_project}. Both systems also retain source generation or repair during task execution. These results expose an unresolved \textbf{\textit{evolution-to-execution problem}}: current CaP methods lack an effective evolution stage, so the previous experiences are not generalizable for runtime execution to benefit from.

Therefore, our paper focuses on the central research question: \textit{how should a CaP system divide evolution-time code from runtime decisions so that it can continue to evolve, transfer across tasks, and still adapt to the current physical state?} Our answer is to move coding to evolution while retaining steerable function calls inside a Reasoning-and-Acting (ReAct) loop at runtime. Code should store physical mechanisms that improve success across tasks. ReAct should retain scene-dependent choices, including action order, function arguments, recovery, and stopping. The interface between them must be strong enough to execute reliable control and flexible enough to expose the choices that may change in a new scene.

Based on this design, we introduce \racap, \textbf{Agentic \underline{R}easoning, \underline{A}cting, and \underline{C}oding \underline{a}s \underline{P}olicies for Evolvable Robot Learning}. As shown in Figure~\ref{fig:method}, RACaP evolves three persistent components before deployment: typed policy application programming interfaces (Policy APIs), the ReAct agent harness, and experience-based long-term memory. Phase 1 uses capability curriculum learning to establish reliable execution, recovery, skill breadth, and orchestration. Phase 2 starts from this foundation and autonomously selects failures, proposes changes, and promotes a candidate only when its paired gains exceed its paired regressions. This two-phase strategy lets the system acquire basic competence before autonomous search, then continue evolving through measured improvements.

At runtime, RACaP freezes source code but preserves adaptive decision making. The Policy APIs encode reusable geometry and contact mechanisms while exposing arguments that ReAct can steer. The ReAct harness maintains task-objective-driven working memory for the current episode and retrieves general lessons from long-term experience memory. Using visual observations, perceptual measurements, and tool reports, ReAct can change its action sequence, API arguments, recovery strategy, and stopping decision without generating new source code. This design supports long-horizon execution while keeping the control loop faster and more stable.

Our experiments validate this evolution-to-execution design across in-domain learning, zero-shot transfer, long-horizon execution, and cross-embodiment evolution. RACaP reaches 54.4\% on LIBERO-90, 45.0\% on zero-shot LIBERO-PRO~\citep{zhou2025liberopro}, and 46.0\% on LIBERO-Long~\citep{libero}. On LIBERO-PRO, it achieves $2.5\times$ the success rate of the strongest CaP baseline with a $1.9\times$ speedup in median policy time. We further distill ReAct decision traces from GPT-5.6 into Qwen3-VL-8B-Instruct through rejection-sampled fine-tuning. The compact model provides a $13.2\times$ per-decision speedup and reduces repeated physical calls by 75\%, showing a practical path toward efficient on-robot execution.

Our contributions are:
\begin{itemize}[leftmargin=*,nosep]
    \item \textbf{Evolution-to-execution formulation.} We separate reusable mechanisms learned through coding from scene-dependent decisions made through runtime function calls. The formulation makes cross-task generalizability and runtime steerability explicit objectives of the learned interface.
    \item \textbf{Two-phase agentic evolution.} We introduce a framework that first builds competence through capability curriculum learning and then improves Policy APIs, the ReAct harness, and long-term experience memory through autonomous paired evolution.
    \item \textbf{Transfer and efficient deployment.} We validate the evolved system on in-domain, zero-shot, long-horizon, and cross-embodiment settings, and show that a distilled ReAct model preserves accepted teacher decisions while reducing inference time and repeated physical calls.
\end{itemize}

\section{Related Work}

\paragraph{Agentic Robot Harnesses.}
Large language models (LLMs) and vision-language models (VLMs) can connect open-ended instructions and visual observations to semantic action plans. SayCan ranks language proposals by skill affordance, while ProgPrompt generates structured task plans~\citep{saycan,progprompt}. Inner Monologue and ReAct then introduce environment feedback into later decisions, and VoxPoser connects language planning to spatial control through 3D value maps~\citep{inner_monologue,react,voxposer}. More recent systems connect the harness to learned visuomotor policies. Harness VLA uses memory-guided agents to stabilize a frozen vision-language-action (VLA) model, while VIMA and OpenVLA provide lower-level multimodal policies~\citep{harness_vla,vima,openvla}. This line progresses from language planning to closed-loop orchestration, but the callable skills usually remain fixed. RACaP extends the harness itself and the Policy APIs together from robot rollouts.

\paragraph{Code as Policy (CaP).}
Coding offers an executable interface for turning a semantic plan into robot behavior. CaP generates programs over perception and control functions, and CaP-X adds visual differencing, execution feedback, and repeated repair~\citep{code_as_policies,cap_x}. VIA treats control as interaction with a visual interface, while RoboClaw and ABot-Claw retain code and state for broader or longer tasks~\citep{via,roboclaw,abot_claw}. Recent progress in CaP focuses on improving later programs from earlier experience. RATS collects reusable code through curiosity-driven play, ASPIRE validates and retrieves program repairs, ENPIRE studies policy improvement on real robots, and GaP organizes multi-agent improvement as a graph of policies~\citep{rats,aspire,enpire,gap}. These methods move from one-shot synthesis toward feedback, persistence, and self-improvement. However, source generation and repair still occur during task execution, and a retained program can mix a reusable mechanism with scene-specific constants and decisions. 

\section{Method}
\label{sec:method}

\subsection{Problem setting and objective}
\label{sec:formulation}

We focus on robotic manipulation tasks. For a task $\tau$, the robot receives an instruction $x$ and observations $o_t$ containing color and depth (RGB-D) images, robot state, and earlier tool reports. The evaluator goal is hidden. The deployable policy has three parts,
\begin{equation}
    \Pi=(\mathcal{A}_{\theta},\pi_{\phi},\mathcal{M}),
    \label{eq:system_parts}
\end{equation}
where $\mathcal{A}_{\theta}$ is the Policy API library with implementation parameters $\theta$, $\pi_{\phi}$ is the runtime ReAct policy defined by harness parameters $\phi$, and $\mathcal{M}$ is long-term experience memory. The runtime state of $\pi_{\phi}$ contains task-objective-driven working memory for the current episode. At each turn, $\pi_{\phi}$ combines this working memory with retrieved lessons from $\mathcal{M}$ to choose a tool and readable arguments. The tool acts and returns a report plus a new image. During deployment, $\pi_{\phi}$ may change the tool or its arguments, but it cannot edit source code.

Let $S(\tau,\mathcal{A}_{\theta},\pi_{\phi},\mathcal{M})\in\{0,1\}$ denote native success on task $\tau$. Let $T(\tau,\mathcal{A}_{\theta},\pi_{\phi},\mathcal{M})$ denote the number of ReAct turns. The development set is $\mathcal{D}_{\mathrm{dev}}$. When $\pi_{\phi}$ and $\mathcal{M}$ are fixed, we choose an API candidate $\theta'$ by its net gain over the current parent $\theta$,
\begin{equation}
\theta^{*}=\arg\max_{\theta'}\sum_{\tau\in\mathcal{D}_{\mathrm{dev}}}
\left[S(\tau,\mathcal{A}_{\theta'},\pi_{\phi},\mathcal{M})-
S(\tau,\mathcal{A}_{\theta},\pi_{\phi},\mathcal{M})\right].
\label{eq:api-objective}
\end{equation}
The parent and candidate use the same tasks, seeds, and budgets. Equation~\ref{eq:api-objective} therefore rewards cross-task generalizability by keeping mechanisms whose paired gains exceed their regressions. Steerability is preserved by exposing typed arguments for choices that may change with the instruction or scene.

When the APIs are fixed, we select ReAct harness parameters $\phi$ and memory $\mathcal{M}$ with a lexicographic~objective,
\begin{equation}
(\phi^{*},\mathcal{M}^{*})=\arg\max_{\phi',\mathcal{M}'}^{\mathrm{lex}}
\left(\sum_{\tau}S(\tau,\mathcal{A}_{\theta^{*}},\pi_{\phi'},\mathcal{M}'),
-\sum_{\tau}T(\tau,\mathcal{A}_{\theta^{*}},\pi_{\phi'},\mathcal{M}')\right).
\label{eq:react-objective}
\end{equation}
We compare success first and turns only when success is tied. This prevents a faster but less successful agent from winning. Together, these objectives suggest that the Policy APIs should reach a balance: it must give ReAct enough flexibility to adapt and recover, while handling low-level skills such as grasping and contact control internally. Appendix~\ref{app:formal} defines the API contract and the complete runtime boundary.

\subsection{Overview}
We use LIBERO-90 as the main development set. Its 90 language-instructed tasks cover transport, spatial relations, insertion, stacking, articulation, controls, and ordered multi-step goals. The system use LIBERO-90 rollouts during both evolution phases. After evolution, we freeze the complete system and test transfer on unseen LIBERO-PRO perturbations. These tests include spatial, goal, and object changes. We also evaluate LIBERO-Long and Robosuite without revealing their native success predicates to the policy. This separation makes LIBERO-90 an in-domain development benchmark and the other suites transfer tests.

Figure~\ref{fig:method} shows how RACaP learns on LIBERO-90 and then freezes the result for evaluation. \textbf{RACaP-Phase 1} follows a fixed capability curriculum that establishes the foundation for self-evolution. Within each stage, the coding agent proposes implementations, but the curriculum determines which capability and task group to improve. \textbf{RACaP-Phase 2} removes this fixed order. The system clusters its current failures, chooses the next capability to target, proposes a change, and validates that change against the current champion. Both phases can update Policy APIs, the ReAct harness, and long-term experience memory. Deployment freezes all three persistent artifacts.

\begin{figure}[tbp]
\centering
\includegraphics[width=0.98\textwidth]{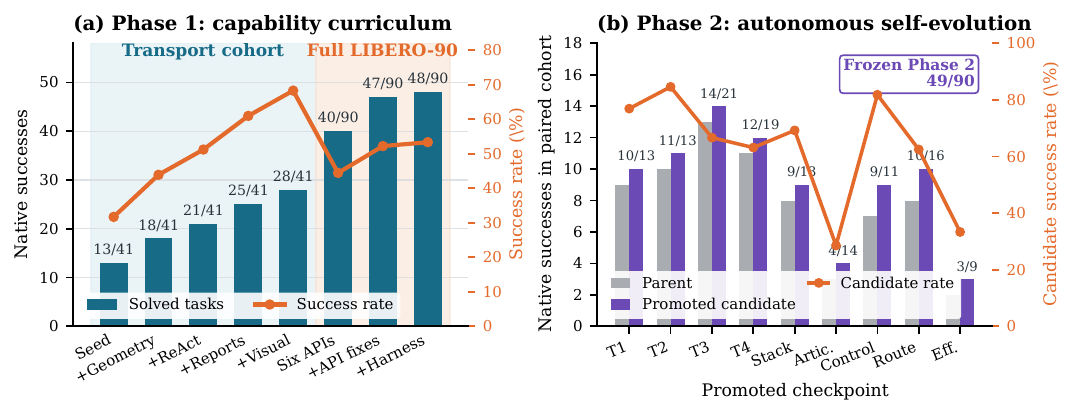}
\caption{\textbf{Capability growth across the two RACaP phases.} (a) Phase 1 first improves a fixed 41-task transport cohort, then expands to all 90 tasks. Bars report solved tasks, and the orange line reports success rate. (b) Phase 2 reports nine winning parent-candidate comparisons. T1--T4 denote transport improvements. The other labels name the tested capability. Each purple bar prints its own task-group denominator.}
\label{fig:phase-progress}
\end{figure}

\subsection{Phase 1: capability curriculum learning}

Autonomous evolution needs a minimum level of competence before its own rollouts become useful learning signals. In preliminary attempts that started from only basic primitives, most trajectories failed before producing informative evidence. One failure could mix perception, grasping, placement, routing, and stopping errors, so the coding agent could not attribute the outcome to one mechanism. As a result, the evolution may fall into poor local minima in which even basic transport never became reliable. We therefore divide evolution into two phases. Phase 1 uses curriculum learning to establish executable and diagnosable behavior before Phase 2 chooses its own improvement path.

The Phase 1 curriculum builds four capability levels across stages S0 to S6, as shown in Figure~\ref{fig:phase-progress}(a) and Appendix Table~\ref{tab:curriculum_unified}. Without access to privileged data like object coordinates, the coding agent learns only from instructions, rollout videos, tool traces, and success signals, ensuring skill stability before long-horizon execution (Appendix~\ref{app:evolution}). On a 41-task transport cohort, \textbf{Execution} establishes basic \texttt{pickplace} (\emph{Seed}) and geometry (\emph{Geometry}), while \textbf{Recovery} adds retries (\emph{ReAct}), failure feedback (\emph{Reports}), and visual correction (\emph{Visual}), raising success from 13/41 (31.7\%) to 28/41 (68.3\%). Across all 90 tasks, \textbf{Breadth} introduces six contact mechanics APIs (\emph{Six APIs}) and stabilizes them (\emph{API fixes}), while \textbf{Orchestration} enhances tool routing, memory, and efficiency (\emph{Harness}), driving full-set performance from 40/90 (44.4\%) to 48/90 (53.3\%).

\subsection{Phase 2: autonomous self-evolution}

Starting from the 48/90 (53.3\%) Phase 1 system, Phase 2 lets an autonomous harness drive further evolution by grouping remaining failures with shared causes, analyzing visual critic summaries of object, gripper, and target movements, and asking a coding agent for single changes across APIs, argument exposures, tool reports, ReAct routing, or memory. To ensure proposed gains outweigh regressions rather than overfitting to isolated rollouts, candidate code is evaluated against the frozen parent system under identical tasks, seeds, and budgets, replacing the parent only when achieving strictly higher native success while archiving ties and regressions (Appendix~\ref{app:evolution}). Over 596 development episodes, the harness generated 32 proposals and tested 23 candidates in simulation, yielding nine promoted wins across capability-specific task groups of varying sizes, covering transport (T1 to T4), stacking, articulation, compact controls, routing, and efficiency (Figure~\ref{fig:phase-progress}(b)). Although in-domain performance on LIBERO-90 increases by only one task to 49/90 (54.4\%), this modest gain masks major progress in zero-shot transfer, runtime stability, and overall robustness, boosting LIBERO-PRO from 32.8\% to 45.0\%, LIBERO-Long from 32.0\% to 46.0\%, and reducing median PRO policy execution time from 454 to 380 seconds.

\subsection{What evolution produces}

\begin{figure}[tbp]
\centering
\includegraphics[width=0.48\linewidth]{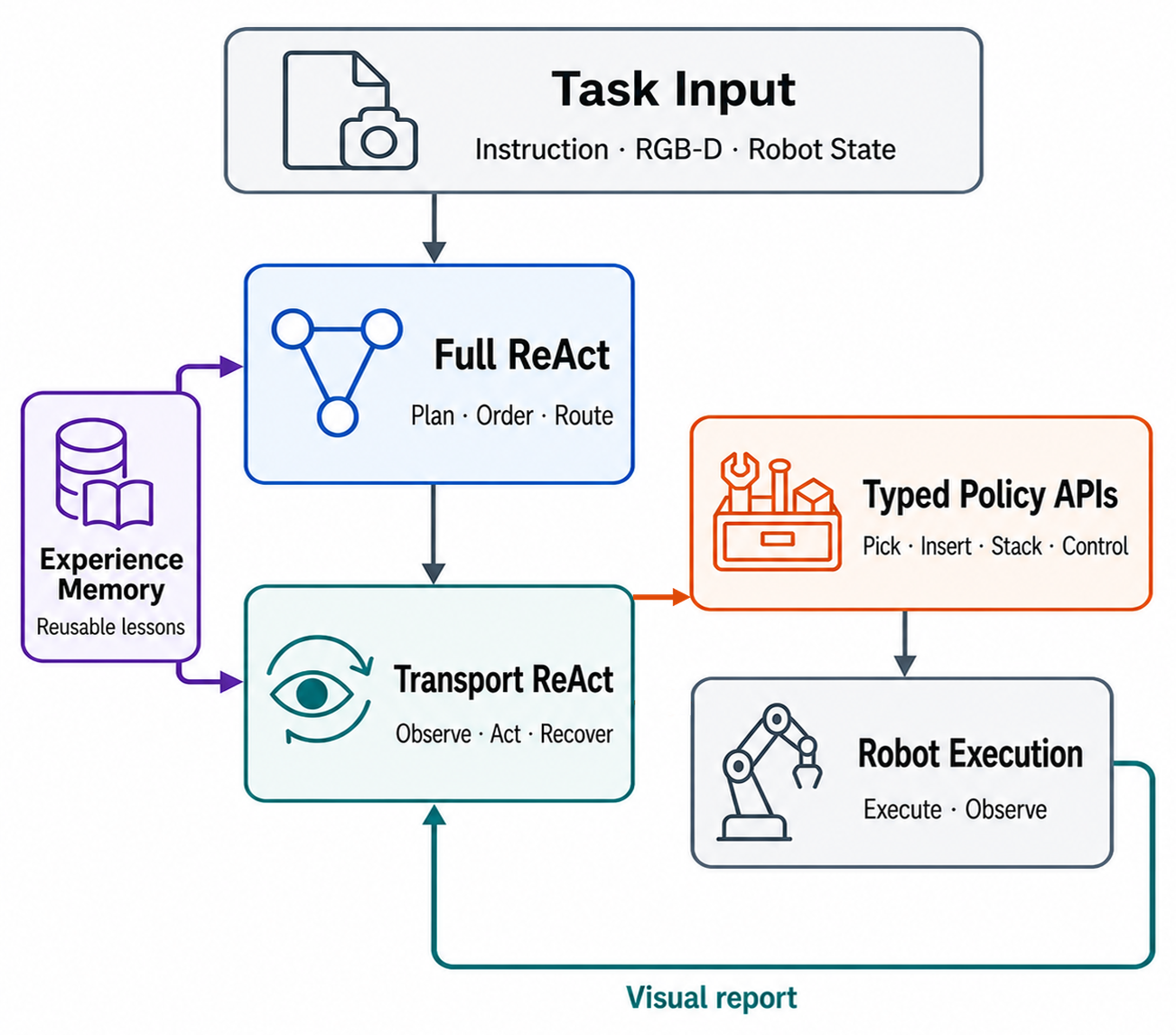}
\caption{\textbf{Frozen RACaP runtime.} Full ReAct routes task-level goals, Transport ReAct handles local object transport, and visual reports close the physical-action loop. Appendix Figure~\ref{fig:react-detail} expands the role-level control and feedback paths.}
\label{fig:runtime-architecture}
\end{figure}

Evolution produces a runtime system with three core components: a Policy API library, a two-tier ReAct agent, and dual-structured memories (Figure~\ref{fig:runtime-architecture}, Appendix Figure~\ref{fig:react-detail}). The Policy API library executes atomic physical operations through \texttt{pickplace}, \texttt{insert}, \texttt{stack}, \texttt{push}, \texttt{articulate}, and \texttt{actuate\_control}, accepting execution parameters including grasp type, clearance, target interpretation, yaw, and contact mode while returning readable status, measurements, and fresh images (Appendix~\ref{app:formal}). Decision-making is divided between Full ReAct, which parses total instructions, orders goals, checks semantic state, and routes across all six APIs, and Transport ReAct, a focused local loop utilizing \texttt{pickplace}, \texttt{insert}, \texttt{push}, \texttt{look}, \texttt{check}, and \texttt{done} (Appendix~\ref{app:react-roles}). Both scopes maintain task-objective-driven working memory to adjust actions upon failure (such as measured offsets triggering nudges or empty grasps switching modes), supported by long-term experience memory that provides general reusable guidance from prior failures without storing privileged task answers.

This structural separation isolates transferable contact geometry within Policy APIs while leaving context-dependent decisions to ReAct. Every physical operation returns control to ReAct, allowing visual tool checks to advise subsequent choices without silently overriding them except on safety or contract violations (Appendix~\ref{app:cases}). Prior to evaluation, the Policy APIs, ReAct harness, long-term experience memory, and selected adapters are strictly frozen, while working memory resets per episode and updates solely through public observations and tool reports. Each execution turn emits a schema-checked call recording arguments, reports, model identity, latency, tokens, and cost, ensuring that RACaP never generates, compiles, or repairs Python code during evaluation.

\subsection{Faster ReAct through rejection-sampled fine-tuning}

To reach faster runtime execution, we distill ReAct decisions from a hosted GPT-5.6-sol teacher into a local Qwen3-VL-8B-Instruct student via rejection-sampled fine-tuning (RFT)~\citep{rrhf,deepseekmath}. The data principle of RFT is retaining teacher rollouts only when they achieve native success, satisfy executable schema constraints, and contain valid visual evidence. Filtering out fluent decisions that lead to incorrect physical states or unexecutable actions yields 54 accepted and 40 rejected decisions, from which we train role-specific LoRA adapters while keeping the base model and vision encoder frozen. RFT increases exact agreement with accepted teacher decisions from 83.3\% to 97.6\% and executable action agreement from 61.9\% to 76.2\%, while reducing repeated physical calls from 16 to 4 across 60 paired held-out episodes. Achieving an average latency of 0.750 seconds per call against 9.895 seconds for the teacher proxy, this $13.2\times$ speedup significantly reduces observation-to-action delay, offering a practical path to local on-robot deployment as a speed showcase rather than a driver of higher native task success (see more details in Appendix~\ref{app:rft-details}).

\section{Experiments}
\label{sec:experiments}

\paragraph{Research questions.}
Our experiments evaluate four core research questions aligned with the system design. \ding{172}~\textbf{Evolution:} do the two learning phases progressively build in-domain competence on LIBERO-90? \ding{173}~\textbf{Generalization:} do frozen artifacts generalize zero-shot to LIBERO-PRO without target-domain feedback? \ding{174}~\textbf{Long-horizon efficiency:} do typed Policy APIs support long-horizon tasks with lower runtime computation than generated code? \ding{175}~\textbf{Deployment breadth:} can the proposed execution boundary support cross-embodiment evolution and compact model distillation? Together, these questions test whether separating evolution from deployment improves both performance and execution efficiency. Full protocols, resource ledgers, confidence intervals, paired tests, and secondary experiments are detailed in Appendix~\ref{app:details}.

\subsection{Experimental setup}

\paragraph{Benchmarks and evaluation boundary.}
LIBERO-90 is the primary development environment and in-domain benchmark, evaluated once per task at seed 0 (since number of the tasks is sufficient, we only use one seed). Zero-shot transfer is evaluated on six LIBERO-PRO suites covering spatial, goal, and object perturbations across 10 tasks and 3 seeds, totaling 180 paired episodes per method. Under zero-shot evaluation, no target-domain rollouts, failure logs, or labels may update code, skills, memory, prompts, or model parameters. Each episode starts in a fresh process and clears all temporary interaction state upon completion. LIBERO-Long evaluates long-horizon execution on 10 tasks across 5 seeds. The seven-object benchmark measures successful placements at 5 minutes, 10 minutes, and 8000 simulator steps. Robosuite evaluates 7 task types across 5 sealed seeds. Final scores are determined by native simulator predicates, which remain hidden from~all~policies. \looseness=-1

\paragraph{Methods and shared runtime.}
We compare five approaches: CaP-X, RATS-base, RATS-90 (after self-play on LIBERO-90), RACaP-Phase 1, and RACaP-Phase 2. CaP-X writes task-specific Python code over fixed non-privileged perception and control APIs. RATS-base runs the official executor without skill reuse or failure memory, while RATS-90 adds a skill library and failure memory trained over 15 LIBERO-90 development rounds. Code-generation baselines may generate and modify temporary source code during an episode. In contrast, both RACaP phases freeze all Policy APIs, the ReAct harness, and memory before evaluation, allowing ReAct to adapt API calls only through visual feedback. All hosted reasoning and visual verification calls use the same GPT-5.5 endpoint. All episodes share an 8000-step horizon, public observations, simulator assets, and native evaluators. \textbf{Note:} To ensure a fair comparison, all methods use GPT-5.5 during evaluation, whereas RACaP uses GPT-5.6 during evolution, which accounts for the Phase 1 performance difference between Figure~\ref{fig:phase-progress}(a) (evolution) and Figure~\ref{fig:main-results}(a) (evaluation). 
Appendix~\ref{app:details} provides complete baseline and budget details. \textit{Controlled RATS comparison.}
Our RATS baselines are evaluated under a standardized setup and fair comparison with other methods. To make a fair comparison with RACaP, our RATS-90 baseline uses 15 development rounds under the shared 596-reset ceiling. Evaluation uses 3 paired seeds per task, the shared GPT-5.5 endpoint (instead of Gemini in their original paper), single continuous episodes, and a maximum of 10 visual code repairs. 

\subsection{In-domain growth and zero-shot transfer}

\begin{figure}[tbp]
\centering
\includegraphics[width=0.97\textwidth]{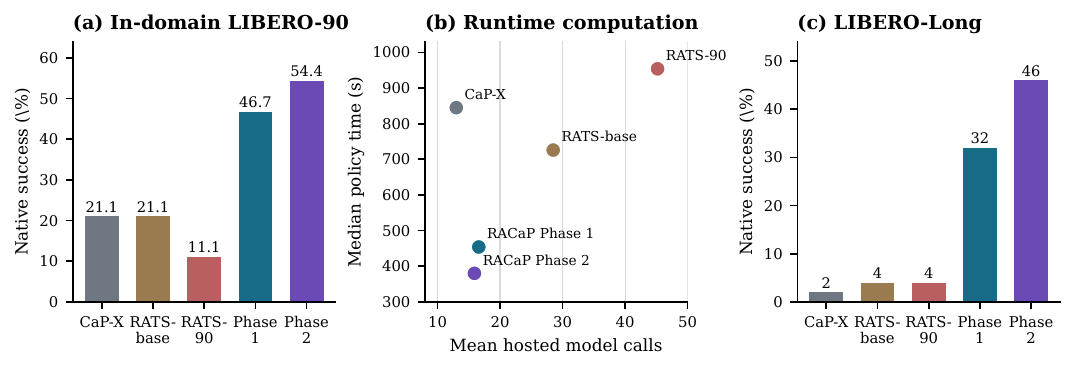}
\caption{\textbf{Main benchmark overview.} (a) In-domain native success on 90 LIBERO-90 tasks (seed 0). (b) Runtime computation across 180 paired LIBERO-PRO episodes, where bottom-left methods indicate fewer model calls and lower policy time. (c) LIBERO-Long results across 10 tasks (5 seeds).}
\label{fig:main-results}
\end{figure}

\noindent\textbf{RACaP builds in-domain competence before transfer.}
Figure~\ref{fig:main-results} summarizes overall performance across LIBERO-90, runtime computation on LIBERO-PRO, and long-horizon tasks. On LIBERO-90, instrumented Phase 1 replay solves 42/90 tasks (46.7\%), while the evolution-time Phase 1 run reaches 48/90 (53.3\%) and the archived Phase 2 artifact solves 49/90 (54.4\%). This confirms that curriculum learning builds the core competence needed for execution, whereas Phase 2 delivers modest additional gains on the~training~distribution. \looseness=-1

\noindent\textbf{Frozen RACaP generalizes beyond its development tasks.}
Table~\ref{tab:libero-pro-breakdown} shows zero-shot success and execution costs across the six LIBERO-PRO perturbation suites. Phase 2 solves 81/180 episodes (45.0\%), compared to 59/180 (32.8\%) for Phase 1 and 15/180 (8.3\%) to 32/180 (17.8\%) for runtime coding CaP baselines. The highest gains occur on object perturbations, reaching 73.3\% for position changes and 60.0\% for task changes, showing that Phase 2 improves generalizable object interfaces rather than memorizing layouts.

Paired analysis highlights significant robustness gains: Phase 2 succeeds on 33 episodes missed by Phase 1, while Phase 1 succeeds on only 11 missed by Phase 2. An exact McNemar test confirms this difference is significant ($p=0.00126$). A similar comparison against RATS-base yields $p=1.41\times10^{-8}$, confirming gains beyond rollout variance.

\begin{table}[tbp]
\caption{\textbf{Frozen-artifact zero-shot LIBERO-PRO native success and runtime.} Success is reported in percent. \emph{Pos.} changes object positions, while \emph{Task} changes the goal and public instruction. Each column contains ten tasks and three seeds, giving 30 episodes. \emph{Avg.} aggregates all 180 episodes. Time is the median policy wall time per episode. Calls and cost are per-episode means.}
\label{tab:libero-pro-breakdown}
\centering
\footnotesize
\setlength{\tabcolsep}{2.6pt}
\begin{tabular}{lrrrrrrrrrr}
\toprule
& \multicolumn{2}{c}{Object} & \multicolumn{2}{c}{Goal} & \multicolumn{2}{c}{Spatial} & & & & \\
\cmidrule(lr){2-3}\cmidrule(lr){4-5}\cmidrule(lr){6-7}
Method & Pos. & Task & Pos. & Task & Pos. & Task & Avg. & Time (s) $\downarrow$ & Calls $\downarrow$ & Cost (\$) $\downarrow$ \\
\midrule
CaP-X & 20.0 & 30.0 & 13.3 & 6.7 & 3.3 & 6.7 & 13.3 & 845 & \textbf{13.0} & 0.68 \\
RATS-base & 36.7 & 26.7 & 16.7 & 10.0 & 13.3 & 3.3 & 17.8 & 726 & 28.5 & 0.91 \\
RATS-90 & 10.0 & 10.0 & 16.7 & 6.7 & 0.0 & 6.7 & 8.3 & 954 & 45.2 & 1.25 \\
\racap-Phase 1 & 26.7 & 50.0 & \textbf{56.7} & 6.7 & 23.3 & 33.3 & 32.8 & 454 & 16.6 & 0.28 \\
\rowcolor{lightgray}
\racap-Phase 2 & \textbf{73.3} & \textbf{60.0} & 46.7 & \textbf{13.3} & \textbf{30.0} & \textbf{46.7} & \textbf{45.0} & \textbf{380} & 15.9 & \textbf{0.24} \\
\bottomrule
\end{tabular}
\end{table}

\begin{figure}[tbp]
\centering
\includegraphics[width=0.93\textwidth]{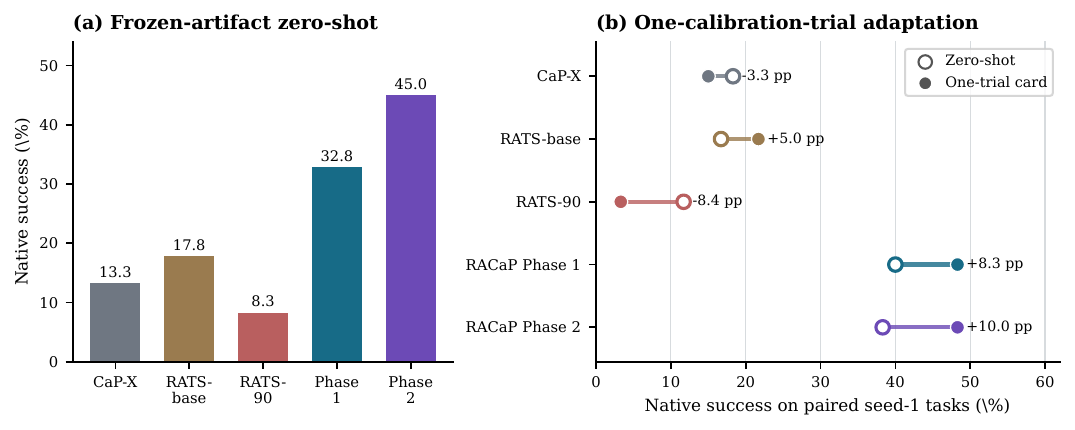}
\caption{\textbf{LIBERO-PRO transfer under two test-time protocols.} (a) Frozen-artifact zero-shot success on all 180 paired episodes. No target-domain rollout updates persistent artifacts. (b) One-calibration-trial adaptation on 60 paired seed-1 episodes per method. Each method receives one calibration rollout for each of the spatial, goal, and object families, stores a small temporary card, and then evaluates the remaining tasks. Open circles show zero-shot success on the same paired subset. Filled circles show success with the card. Appendix Table~\ref{tab:one-trial-app} gives counts and paired tests.}
\label{fig:pro-transfer}
\end{figure}

\noindent\textbf{Persistent code helps only when its interfaces remain compatible.}
Phase 2 achieves the highest LIBERO-PRO performance while averaging a median policy time of 380 seconds and 15.9 API calls. In contrast, RATS-90 requires 45.2 calls and 954 seconds, dropping from 17.8\% (RATS-base) to 8.3\%. Audits trace these failures to interface drift between saved helper functions and newly generated caller scripts, including mismatched keys, data types, and function arguments. This highlights that persistent code reuse requires strict typed interfaces and regression testing; without them, self-play can introduce API mismatches and raise repair costs. Appendix~\ref{app:details} provides the complete RATS-90 audit.\looseness=-1

\noindent\textbf{One calibration trial reveals adaptation capacity.}
Figure~\ref{fig:pro-transfer} contrasts zero-shot transfer with a one-trial adaptation diagnostic. A single calibration trial per task family improves Phase 1 by 8.3 percentage points and Phase 2 by 10.0 points on the seed-1 subset. Code-generation baselines show mixed or negative responses.

\noindent\textbf{Long-horizon gains come from composition, not shorter traces.}
On LIBERO-Long, Phase 2 solves 23/50 episodes (46.0\%) and Phase 1 solves 16/50 (32.0\%), whereas code-generation baselines solve only 2.0\% to 4.0\%. Gains are concentrated in subtasks like \texttt{On} (40.0\% to 65.0\%) and \texttt{Turnon} (20.0\% to 50.0\%). Since mean ReAct turns increase from 8.22 to 8.72, improvements stem from completing complex subgoal sequences rather than shortening decision traces. McNemar testing yields $p=0.118$, and \texttt{Close} remains at 0.0\% (Appendix Table~\ref{tab:long-predicate-app}). On the seven-object benchmark, Phase 2 completes a median of 3.0 objects at 10 minutes, though no method completes all seven (Appendix Figure~\ref{fig:secondary-results}).

\subsection{Cross embodiment evolution and compact ReAct}

\noindent\textbf{Cross-embodiment transfer remains difficult, but the interface can evolve.}
Robosuite tests cross-embodiment transfer by changing both robot hardware and control backends. Without target-domain evolution, frozen Phase 2 solves 6/35 episodes (17.1\%) and CaP-X solves 9/35 (25.7\%). After domain evolution, RACaP-RS and RATS-RS both achieve 11/35 success (31.4\%), but RACaP-RS requires only 32.9 seconds and \$0.089 per episode versus 90.7 seconds and \$0.159 for RATS-RS. Both solve 1/10 held-out task types (10.0\%), demonstrating that typed interfaces facilitate efficient target-domain adaptation rather than direct zero-shot embodiment transfer (Appendix Table~\ref{tab:robosuite-app}).

\begin{figure}[tbp]
\centering
\includegraphics[width=0.88\textwidth]{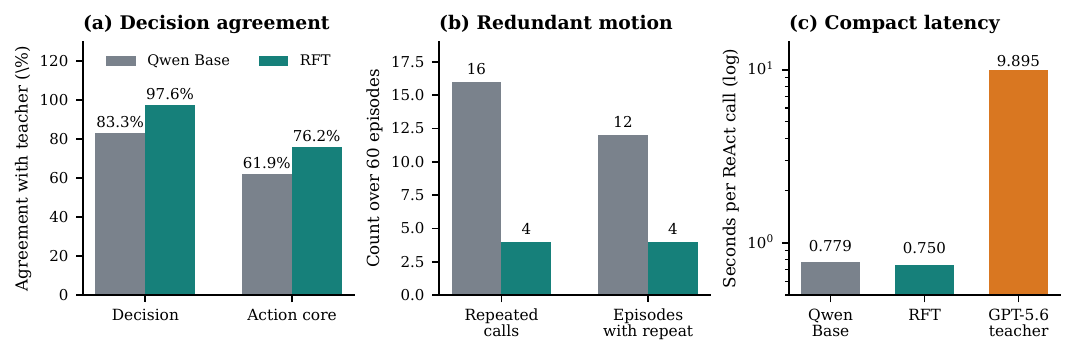}
\caption{\textbf{Compact ReAct diagnostics.} RFT approaches accepted teacher decisions, reduces redundant motion, and retains compact-model latency. Teacher latency is an unpaired reference.}
\label{fig:rft-main}
\end{figure}

\noindent\textbf{A compact ReAct student preserves teacher decisions with lower latency.}
Figure~\ref{fig:rft-main} shows that RFT increases exact agreement with teacher decisions from 83.3\% to 97.6\% and action-schema agreement from 61.9\% to 76.2\%. Across 60 paired evaluation episodes, repeated physical action calls decrease from 16 to 4. The compact student averages 0.750 seconds per call versus 9.895 seconds for the hosted teacher model. Combining typed Policy APIs with RFT eliminates runtime coding overhead while reducing ReAct latency and redundant actions, offering a practical pipeline for local on-robot deployment (Appendix~\ref{app:rft-details}).

\section{Limitations and Broader Impact}

Freezing source code at deployment restricts RACaP to generalizing within its Policy API closure rather than inventing new physical skills online. Future systems could maintain this fast frozen path for standard execution while routing novel failures to a sandboxed coding pipeline that promotes updated code. Despite these limits,  RACaP shows how a CaP system can move from experimental code generation toward an evolvable deployment stack. It will push CaP methods a step further to a more steerable, transferrable, efficient embodied intelligence.

\section{Conclusion}

Addressing the evolution-to-execution trade-off in Code as Policy systems, RACaP decouples reusable physical mechanisms learned as code during offline evolution from scene-dependent choices managed at runtime. By evolving steerable Policy APIs, a ReAct harness, and long-term experience memory through curriculum learning and paired self-play, RACaP freezes source code at deployment while adapting function calls through working memory and visual feedback. Across in-domain, zero-shot, and long-horizon LIBERO benchmarks, this design improves policy transfer while significantly reducing online computation compared to runtime code generation. Furthermore, Robosuite experiments demonstrate efficient cross-embodiment adaptation, while compact ReAct distillation provides a viable pathway for real-time local execution. Ultimately, separating persistent code structures from dynamic agent decisions enables robot policies to continuously evolve without sacrificing transfer ability or execution efficiency.

\bibliography{references}
\bibliographystyle{plainnat}

\clearpage
\appendix

\begin{center}
  \Large\bfseries Appendix
\end{center}

\section{Formal Objective}
\label{app:formal}

This section defines the interface optimized by Equations~\ref{eq:api-objective} and~\ref{eq:react-objective}, explicitizing the responsibility boundary between evolution and deployment. Table~\ref{tab:boundary} details the architectural ownership across system layers.

\begin{table}[htbp]
\caption{\textbf{Responsibility boundary across system layers.}}
\label{tab:boundary}
\centering
\small
\begin{tabular}{P{0.22\columnwidth}P{0.36\columnwidth}P{0.34\columnwidth}}
\toprule
Layer & Owns & Excludes \\
\midrule
Policy API & Local contact geometry, motion execution, physical measurements, failure reporting & Whole-task instruction, causal goal ordering, semantic state checks \\
Runtime ReAct & Task decomposition, tool and argument selection, failure recovery, stopping logic & Source code modifications, privileged evaluator state, hidden task branches \\
Working Memory & Current instruction, remaining subgoals, recent API calls, visual outcomes & Cross-episode learning, source code edits, simulator state \\
Experience Memory & Reusable action prerequisites, recovery lessons, API capability boundaries & Task answers, absolute scene coordinates, evaluator state \\
Evolution Harness & Rollout analysis, code candidate generation, paired regression, version control & Deployment action choice, simulator state in policy input \\
\bottomrule
\end{tabular}
\end{table}

\begin{figure}[tbp]
\centering
\includegraphics[width=0.90\textwidth]{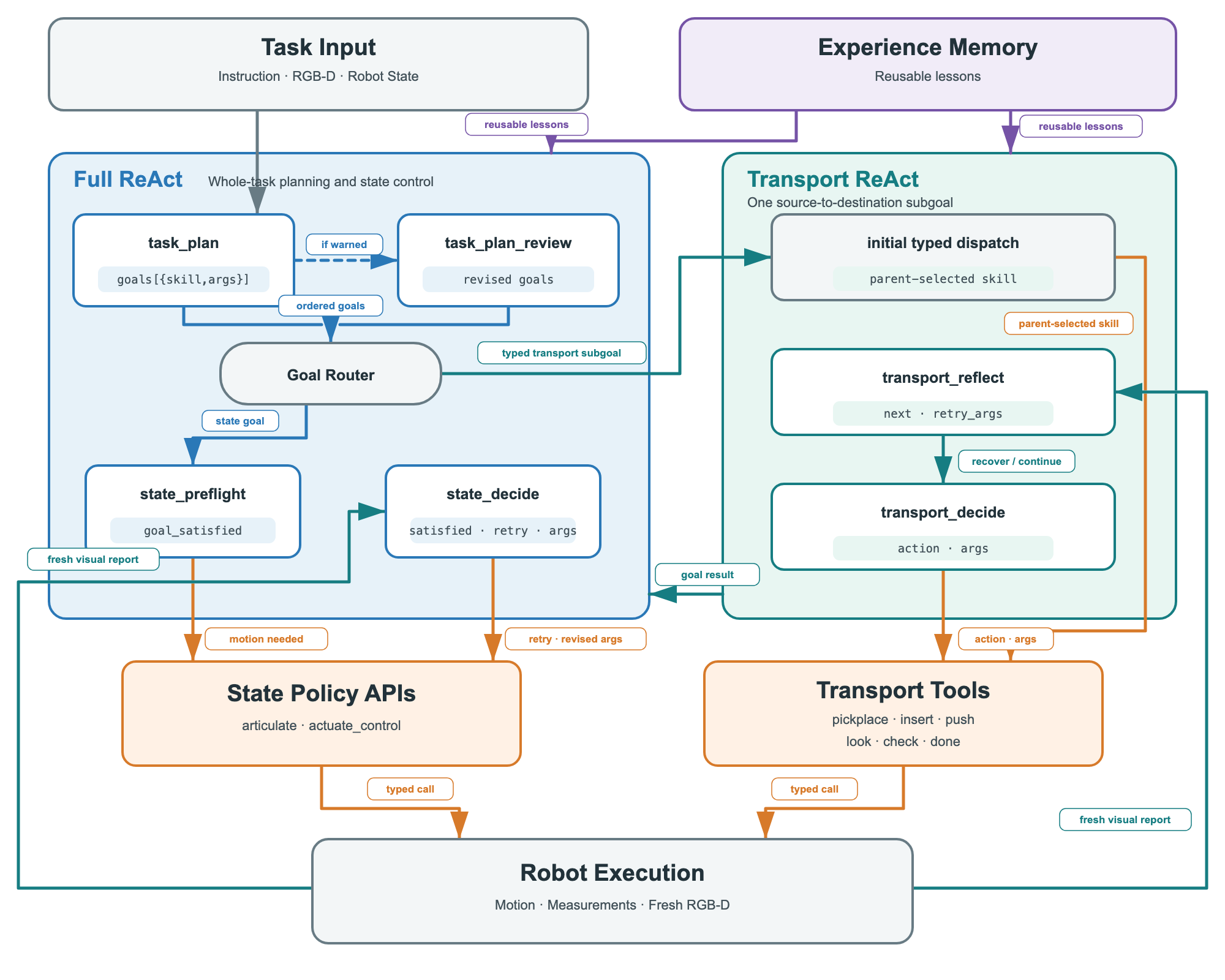}
\caption{\textbf{Detailed role-level RACaP runtime architecture.} Full ReAct manages overall task decomposition and mechanism-state control. Transport ReAct handles source-to-destination subgoals. Orange paths trigger physical tools, with post-action judgments receiving measurements and fresh RGB-D images only after Robot Execution. Purple paths supply read-only experience memory to both decision scopes.}
\label{fig:react-detail}
\end{figure}

\subsection{ReAct Scopes and Roles}
\label{app:react-roles}

RACaP structures runtime decisions into two hierarchical scopes rather than six independent agents. Full ReAct parses the complete instruction and establishes causal task ordering. Transport ReAct operates as a child controller dedicated to single source-to-destination subgoals. Both scopes maintain task-objective-driven working memory to track goal progress, augmented by read-only lessons retrieved from long-term experience memory. Decision points are governed by typed prompt \emph{roles}, which define exact visual inputs and required JSON output fields without directly executing environment actions.

Table~\ref{tab:react-role-contracts} outlines the six runtime prompt contracts. Full ReAct utilizes four roles: \texttt{task\_plan} formulates typed subgoals, \texttt{task\_plan\_review} runs conditionally upon syntax or ordering warnings, while \texttt{state\_preflight} and \texttt{state\_decide} govern articulated mechanism and appliance control before and after physical execution. Transport ReAct employs \texttt{transport\_decide} for local tool selection and \texttt{transport\_reflect} for post-motion visual diagnosis using paired before-and-after images.

Figure~\ref{fig:react-detail} illustrates the underlying control flow. Every physical API execution places Robot Execution between action dispatch and visual judgment. Consequently, API status reports, physical measurements, and fresh RGB-D frames reflect true physical outcomes, which are subsequently logged into working memory for active closed-loop control.

\begin{table}[tbp]
\caption{\textbf{Typed contracts for the six runtime ReAct roles.} Output fields list action-bearing JSON schemas.}
\label{tab:react-role-contracts}
\centering
\footnotesize
{\setlength{\tabcolsep}{5pt}
\begin{tabular}{P{0.08\textwidth}P{0.20\textwidth}P{0.16\textwidth}P{0.22\textwidth}P{0.23\textwidth}}
\toprule
Scope & Role & Trigger Condition & Public Observation Input & Action-Bearing JSON Output \\
\midrule
Full ReAct & \texttt{task\_plan} & Instruction start or replanning trigger & Task instruction, current RGB-D image, skill schemas, Full memory & Ordered \texttt{goals} with specified \texttt{skill} and typed arguments \\
Full ReAct & \texttt{task\_plan\_review} & Syntax or ordering warning from checker & Instruction, candidate plan, checker warnings, RGB-D image, Full memory & Revised ordered \texttt{goals} following identical schema \\
Full ReAct & \texttt{state\_preflight} & Prior to articulation or control motion & Typed state goal, RGB-D image, advisory state report, Full memory & \texttt{goal\_satisfied} boolean flag \\
Full ReAct & \texttt{state\_decide} & Following articulation or control motion & State goal, attempted call, tool report, before/after/home images, visual crop & \texttt{goal\_satisfied}, \texttt{retry} flag, and updated mechanism arguments \\
Transport ReAct & \texttt{transport\_decide} & Local loop initialization or recovery step & Single transport goal, current image, local call history, gripper state, schemas & Target \texttt{action} and typed \texttt{args} dictionary \\
Transport ReAct & \texttt{transport\_reflect} & Following transport execution & Instruction, attempted call, tool reports, gripper state, before/after images & Visual outcome \texttt{next} action and optional \texttt{retry\_args} \\
\bottomrule
\end{tabular}
}
\end{table}

\Needspace{14\baselineskip}
\paragraph{JSON Contract Specifications.}
Full-task planning roles return structured goal sequences referencing Policy API arguments without exposing raw object coordinates:
\begin{Verbatim}[samepage=true,breaklines=true,breakanywhere=true,fontsize=\small]
{
  "goals": [
    {"skill": "pickplace|insert", "pick": "...", "place": "..."},
    {"skill": "stack", "pick": "...", "support": "..."},
    {"skill": "articulate", "target": "...", "part": "...",
     "goal": "open|closed"},
    {"skill": "actuate_control", "target": "...", "control_hint": "...",
     "goal": "on|off"}
  ],
  "reason": "..."
}
\end{Verbatim}

Mechanism-state roles prioritize visual evidence over advisory sensor readings. \texttt{state\_preflight} skips motion if the goal state is already achieved:
\begin{Verbatim}[samepage=true,breaklines=true,breakanywhere=true,fontsize=\small]
{"goal_satisfied": true, "what_you_see": "...", "reason": "..."}
\end{Verbatim}
Following physical execution, \texttt{state\_decide} evaluates visual differences across before, after, and home-cleared~views:
\begin{Verbatim}[samepage=true,breaklines=true,breakanywhere=true,fontsize=\small]
{"goal_satisfied": true, "retry": false, "what_you_see": "...",
 "args": {"mechanism": "auto", "contact_strategy": "...", 
 "amount": null}, 
 "reason": "..."}
\end{Verbatim}

Transport subgoals manage local object transfer through \texttt{transport\_decide} and \texttt{transport\_reflect}. The reflection step analyzes visual outcomes to output structured recovery directives:
\begin{Verbatim}[samepage=true,breaklines=true,breakanywhere=true,fontsize=\small]
{"what_happened": "...", "looks_placed": true,
 "next": "check", "retry_args": {}, "reason": "..."}
\end{Verbatim}

\section{Evolution Details}
\label{app:evolution}

Table~\ref{tab:curriculum_unified} details the unified Phase 1 capability curriculum across stages S0 to S6, defining the task distributions, visual evidence, and corresponding system modifications.

\begin{table}[tbp]
\caption{\textbf{Unified Phase 1 capability curriculum and evolutionary modifications.}}
\label{tab:curriculum_unified}
\centering
\small
\begin{tabular}{P{0.06\textwidth}P{0.17\textwidth}P{0.37\textwidth}P{0.29\textwidth}}
\toprule
Stage & Capability & Benchmark Tasks and Visual Evidence & Primary System Artifact Modified \\
\midrule
S0 & Direct Transport & Moving visible objects to open targets or containers & \texttt{pickplace} geometry routines \\
S1 & Transport ReAct & Recovery from empty grasps and offset placements & Local ReAct prompts and retry parameters \\
S2 & Bounded Placement & Distinguishing target identity and valid interior regions & Destination grounding and \texttt{check} reports \\
S3a & Insertion & Insertion across books, mugs, caddies, and shelves & \texttt{insert} footprint and orientation control \\
S3b & Stacking & Aligning support surfaces and hollow nesting geometries & \texttt{stack} support stability verifications \\
S3c & Articulation & Door, drawer, and cabinet opening along line/arc paths & \texttt{articulate} trajectory contact paths \\
S3d & Compact Controls & Operating stove knobs, buttons, and short switches & \texttt{actuate\_control} parameterization \\
S4 & Tool Routing & Selecting appropriate Policy APIs for single goals & Full ReAct tool selection schema \\
S5 & Causal Composition & Sequenced execution involving open, place, and close & Task planner and prerequisite memory \\
S6 & Efficiency & Eliminating redundant checks and ineffective retries & Harness execution budgets and scoped memory \\
\bottomrule
\end{tabular}
\end{table}

The code repository isolates core modular components across specific directories: Policy APIs reside in \texttt{racap/policy\_api/}, Full and Transport ReAct agents are implemented in \texttt{racap/agent/full\_react.py} and \texttt{racap/agent/react.py}, tool schemas are specified in \texttt{racap/agent/tools.py}, frozen experience memory is stored in \texttt{evolution/champion\_code/memory/}, and the evolutionary harness logic is maintained within~\texttt{evolution/harness/}. \looseness=-1

In Phase 2 self-play, code proposals include predicted gains, regression risks, and explicit rejection conditions. Visual critics align before and after frames with API calls. Proposals pass syntax compilation and basic test suites before entering simulator validation, where promotion requires a strictly positive success delta over the parent system.

\section{Qualitative Analysis}
\label{app:cases}

\subsection{Closed-Loop Visual Recovery}

Table~\ref{tab:recovery-trace} illustrates a representative zero-shot LIBERO-PRO recovery trace on \texttt{libero\_spatial\_task/6/seed0} (transferring a bowl from a stove to a plate). Unlike baseline systems that repeat failing trajectories, RACaP Phase 2 uses visual checks to iteratively adjust subsequent execution parameters, transitioning to a minor push upon reducing spatial error to 3.2 cm.

\begin{table}[tbp]
\caption{\textbf{Condensed Phase 2 visual recovery trace on \texttt{libero\_spatial\_task/6/seed0}.}}
\label{tab:recovery-trace}
\centering
\small
\begin{tabular}{P{0.08\textwidth}P{0.20\textwidth}P{0.62\textwidth}}
\toprule
Turn & API Action & Observation Feedback and Reactive Decision \\
\midrule
1--2 & \texttt{pickplace}, \texttt{check} & Measured offset is 19.5 cm from target center. Initiate retry with measured nudge. \\
3--4 & \texttt{pickplace}, \texttt{check} & Error decreases to 6.3 cm. Adjust nudge direction rather than repeating past trajectory. \\
5--6 & \texttt{pickplace}, \texttt{check} & Error decreases to 3.2 cm. Switch to local \texttt{push} instead of full regrasp. \\
7--8 & \texttt{push}, \texttt{check} & Grounding verification yields 2.0 cm alignment, satisfying placement threshold. \\
9 & \texttt{done} & ReAct terminates execution based on visual validation. Native evaluator confirms success. \\
\bottomrule
\end{tabular}
\end{table}

\subsection{Interface Abstraction vs. Code Generation}

Code-generation approaches like CaP-X construct raw Python scripts upon receiving instructions, segmenting objects and hardcoding explicit spatial coordinates (e.g., \texttt{place\_pos = [basket\_x, basket\_y, basket\_top + 0.16]}) alongside low-level \texttt{goto\_pose} directives. Subsequent repairs generate entirely new code blocks, creating vulnerability to syntax errors and state mismatches.

Conversely, RACaP encapsulates contact geometry within tested Policy APIs, issuing structured calls such as \texttt{pickplace(pick=..., place=..., nudge=...)} and \texttt{push(nudge=...)}. This abstraction isolates shape reasoning, clearance calculations, and motion generation within versioned execution routines, allowing ReAct to adapt grasp strategies and corrective nudges purely through typed parameter adjustments.

\section{Experimental Details}
\label{app:details}

\paragraph{Evaluation Protocols.}
Evaluations span 90 tasks in LIBERO-90 (seed 0), 180 episodes in LIBERO-PRO (6 suites, 10 tasks, seeds 0--2), and 50 episodes in LIBERO-Long (10 tasks, seeds 0--4). Standard episodes are allocated one reset, an 8000-step simulator limit, and a 1000-second policy wall-clock limit. Code-generation baselines are capped at 10 visual feedback repairs per episode. All hosted LLM calls utilize GPT-5.5 via a unified VAPI endpoint at zero temperature. Statistical reporting includes Wilson 95\% confidence intervals, exact paired McNemar tests, and Holm corrections.

\paragraph{RATS-90 Regression Audit.}
RATS-90 success drops to 8.3\% (15/180) on LIBERO-PRO compared to 17.8\% (32/180) for RATS-base, with the largest regression occurring on spatial position perturbations (0/30 vs 4/30). Detailed trace audits reveal that RATS-90 fails within its learned \texttt{locate\_verified\_object\_3d} function prior to physical movement, stemming from contract drift between caller scripts and saved functions (such as missing \texttt{agentview} keys, unexpected array formats, or unhandled arguments). These interface mismatches account for 181 out of 321 total RATS-90 failures across the evaluation grid, demonstrating that unconstrained self-play without typed interfaces can aggravate failure rates.

\paragraph{Development Resources.}
Table~\ref{tab:development-resources-app} details the development resource consumption during Phase 2 evolution and RATS self-play iterations under equalized simulator reset ceilings.

\begin{table}[htbp]
\caption{\textbf{Audited development resource consumption.}}
\label{tab:development-resources-app}
\centering
\small
\begin{tabular}{lrrrr}
\toprule
Lineage & Iteration Units & Simulator Episodes & Hosted API Calls & Total Cost (\$) \\
\midrule
RACaP-Phase 2 & 32 proposals & 596 & n/a & n/a \\
RATS (Committed) & 15 rounds & 206 & 2,474 & 62.55 \\
RATS (Valid Append-Only) & 15 rounds & 252 & 3,056 & 74.43 \\
\bottomrule
\end{tabular}
\end{table}

\paragraph{Extended Benchmark Diagnostics.}
Table~\ref{tab:long-predicate-app} breaks down performance across predicate types on LIBERO-Long. Table~\ref{tab:one-trial-app} presents results for the one-trial adaptation diagnostic on LIBERO-PRO, where single family-level experience cards are evaluated on seed-1 tasks. Table~\ref{tab:robosuite-app} details performance on Robosuite across 7 task types and 5 seeds.

\begin{table}[htbp]
\caption{\textbf{LIBERO-Long predicate breakdown (50 total episodes).}}
\label{tab:long-predicate-app}
\centering
\small
\begin{tabular}{lrrrrrr}
\toprule
Method & Native Success (\%) & \texttt{In} (\%) & \texttt{On} (\%) & \texttt{Turnon} (\%) & \texttt{Close} (\%) & ReAct Turns \\
\midrule
RACaP-Phase 1 & 32.0 & 26.7 & 40.0 & 20.0 & 0.0 & 8.22 \\
RACaP-Phase 2 & \textbf{46.0} & \textbf{33.3} & \textbf{65.0} & \textbf{50.0} & 0.0 & 8.72 \\
\bottomrule
\end{tabular}
\end{table}

\begin{table}[htbp]
\caption{\textbf{One-trial adaptation performance on 60 paired LIBERO-PRO seed-1 episodes.}}
\label{tab:one-trial-app}
\centering
\small
\begin{tabular}{lrrr}
\toprule
Method & Zero-Shot Success (\%) & One-Trial Success (\%) & Raw $p$-value \\
\midrule
CaP-X & 18.3 & 15.0 & 0.791 \\
RATS-base & 16.7 & 21.7 & 0.629 \\
RATS-90 & 11.7 & 3.3 & 0.180 \\
RACaP-Phase 1 & 40.0 & \textbf{48.3} & 0.302 \\
RACaP-Phase 2 & 38.3 & \textbf{48.3} & 0.146 \\
\bottomrule
\end{tabular}
\end{table}

\begin{table}[htbp]
\caption{\textbf{Sealed Robosuite evaluation across 35 episodes (7 task types, 5 seeds).}}
\label{tab:robosuite-app}
\centering
\small
\begin{tabular}{lrrrr}
\toprule
Method & Native Success (\%) & Median Policy Time (s) & API Calls & Cost (\$) \\
\midrule
CaP-X & 25.7 & 82.4 & 2.1 & 0.111 \\
RATS-90 & 17.1 & 104.3 & 3.0 & 0.169 \\
RACaP-Phase 2 frozen & 17.1 & 44.8 & 7.8 & 0.106 \\
RATS-RS & \textbf{31.4} & 90.7 & 3.0 & 0.159 \\
RACaP-RS & \textbf{31.4} & \textbf{32.9} & 6.7 & \textbf{0.089} \\
\bottomrule
\end{tabular}
\end{table}

\subsection{Rejection-Sampled Fine-Tuning}
\label{app:rft-details}

Rejection-sampled fine-tuning (RFT) distills ReAct decisions from a hosted GPT-5.6-sol teacher into a local Qwen3-VL-8B-Instruct student. Teacher collection leverages 24 selected LIBERO-90 training trajectories. The dataset filter enforces native execution success, validates role-specific output schemas, and verifies image reference integrity, retaining 54 decisions while rejecting 40.

Training optimizes role-specific LoRA adapters while keeping base model weights and the vision encoder frozen. Loss is applied exclusively to teacher response tokens. Evaluation across 60 paired held-out episodes demonstrates that RFT reduces repeated physical calls from 16 to 4, decreasing the proportion of episodes containing redundant calls from 20.0\% (12/60) to 6.7\% (4/60). The distilled student achieves an average latency of 0.750 seconds per ReAct decision, compared to 9.895 seconds for the hosted teacher proxy.

\begin{figure}[tbp]
\centering
\begin{minipage}[t]{0.49\textwidth}
\centering
\includegraphics[width=\linewidth]{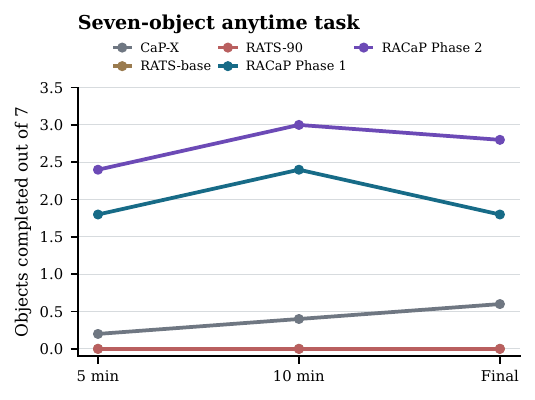}
\end{minipage}\hfill
\begin{minipage}[t]{0.49\textwidth}
\centering
\includegraphics[width=\linewidth]{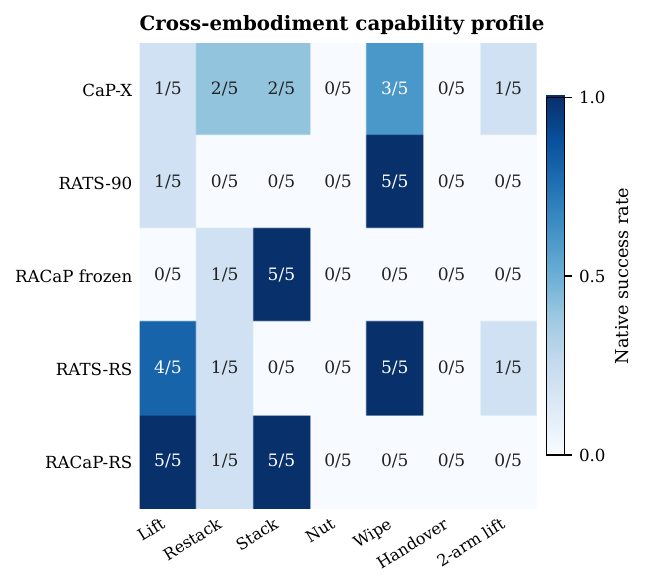}
\end{minipage}
\caption{\textbf{Left:} cumulative task completion on the seven-object benchmark over time. \textbf{Right: }per-task performance breakdown across Robosuite domains.}
\label{fig:secondary-results}
\end{figure}

\end{document}